\def\year{2022}\relax
\documentclass[letterpaper]{article} 
\usepackage{aaai22}  
\usepackage{times}  
\usepackage{helvet}  
\usepackage{courier}  
\usepackage[hyphens]{url}  
\usepackage{graphicx} 
\usepackage{natbib}  
\usepackage{caption} 
\DeclareCaptionStyle{ruled}{labelfont=normalfont,labelsep=colon,strut=off} 
\usepackage{amsmath}
\usepackage{amssymb}
\usepackage{algorithm}
\usepackage{algorithmic,esint}
\usepackage{multirow}
\usepackage{booktabs} 
\usepackage{enumitem}
\usepackage{comment}
\usepackage{bm} 
\usepackage{subfig}
\usepackage{newfloat}
\usepackage{listings}
\floatstyle{ruled}
\newfloat{listing}{tb}{lst}{}
\floatname{listing}{Listing}
\title{Learning to Solve Travelling Salesman Problem \\ with Hardness-Adaptive Curriculum}
\author{
    Zeyang Zhang,\;
    Ziwei Zhang,\;
    Xin Wang\thanks{Corresponding authors},\;
    Wenwu Zhu\footnotemark[1] \\
}
\affiliations{
    Tsinghua University \\
    zy-zhang20@mails.tsinghua.edu.cn, zwzhang@tsinghua.edu.cn \\
    xin\_wang@tsinghua.edu.cn, wwzhu@tsinghua.edu.cn

}

\begin{document}

\maketitle

\begin{abstract}
\begin{quote}
Various neural network models have been proposed to tackle combinatorial optimization problems such as the travelling salesman problem (TSP). Existing learning-based TSP methods adopt a simple setting that the training and testing data are independent and identically distributed. However, the existing literature fails to solve TSP instances when training and testing data have different distributions.
Concretely, we find that different training and testing distribution will result in more difficult TSP instances, i.e., the solution obtained by the model has a large gap from the optimal solution.
To tackle this problem, in this work, we study learning-based TSP methods when training and testing data have different distributions using adaptive-hardness, i.e., how difficult a TSP instance can be for a solver. 
This problem is challenging because it is non-trivial to (1) define hardness measurement quantitatively; (2) efficiently and continuously generate sufficiently hard TSP instances upon model training; (3) fully utilize instances with different levels of hardness to learn a more powerful TSP solver.
To solve these challenges, we first propose a principled hardness measurement to quantify the hardness of TSP instances. Then, we propose a hardness-adaptive generator to generate instances with different hardness. We further propose a curriculum learner fully utilizing these instances to train the TSP solver. Experiments show that our hardness-adaptive generator can generate instances ten times harder than the existing methods, and our proposed method achieves significant improvement over state-of-the-art models in terms of the optimality gap.
\end{quote}
\end{abstract}

\section{Introduction}
The travelling salesman problem (TSP), as one important NP-hard problem, serves as 
a common benchmark for evaluating combinatorial optimization (CO) algorithms that have many practical
real-world applications.
As a trade-off between computational costs and solution qualities, a collection of approximate solvers and heuristics have been studied~\cite{TSP-heuristic,lkh3}.
On the other hand, 
there is a recent advent of using machine learning to facilitate solving the NP-hard travelling salesman problem (TSP) in an end-to-end fashion~\cite{Ptr-TSP15,GAT-RL-TSP19,NE-RL-TSP17,Ptr-RL-TSP17,POMO-RL-TSP20,AM-Ptr-RL-TSP18,GNN-Ptr-RL-TSP19,TSP-transformer21}.
However, the existing methods independently sample training and testing data from the same distribution, i.e., the i.i.d. setting. More concretely, most methods directly adopt a uniform sampling within the unit square to generate TSP instances. Therefore, though these existing methods show reasonably good performance in such a setting, it is unclear whether the solver trained under the existing i.i.d. setting can actually solve the TSP problem by modeling and capturing the underlying relationships between instances and solutions or simply memorizing training instances. If the latter, the existing TSP solver will fail to handle when training and testing TSP have various distributions.

\begin{figure}
    \centering    \includegraphics[width=0.95\columnwidth]{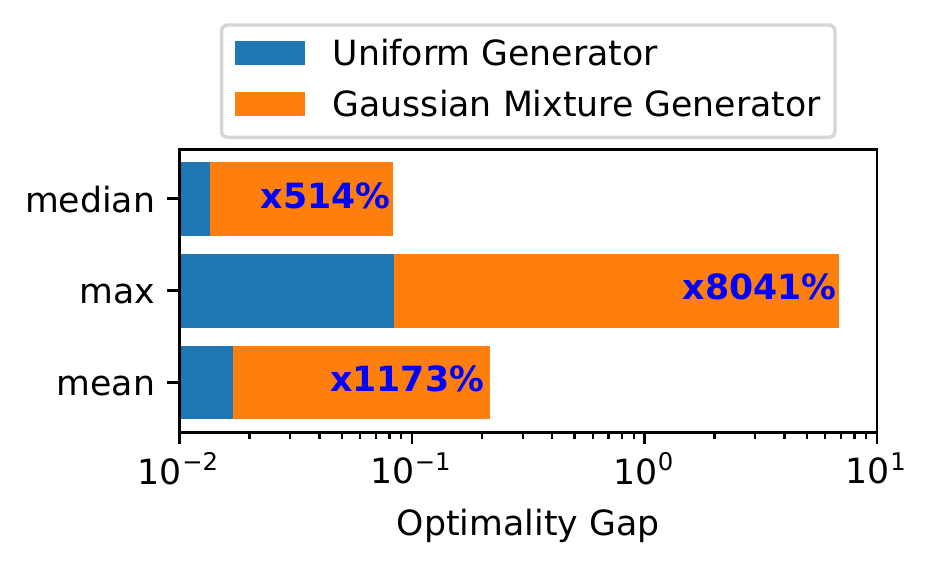}
    \caption{The optimality gap of an existing TSP solver trained on uniform samples and 
    tested on Gaussian mixture samples. The maximum optimality gap 
    grows 8041\% times larger, showing the weakness of the existing models when training and testing data have different distributions.}
    \label{fig:weak measure}
\end{figure}

To answer this question, we first conduct some preliminary studies for the existing TSP solvers. Specifically, we replace the uniform samples in the testing phase with Gaussian mixture samples, which are significantly harder TSP instances whose solutions obtained from a TSP solver have a larger gap from
the optimal solution (for more details, please refer to Section~\ref{sec:prelim}). The results are shown in Figure~\ref{fig:weak measure}. The figure shows that indeed as we speculate, the existing TSP solvers fail to generalize to this more challenging setting where training and testing data have different distributions. As the distribution changes, the model performance degrades several times with respect to all metrics. The results clearly demonstrate that the existing TSP solvers have serious deficiencies when training and testing data have different distributions, and these instances from different distributions are much more difficult.

To solve this problem, in this paper, we study learning-based TSP methods when training and testing data have different distributions using adaptive-hardness.
However, there exist several challenges.
\begin{itemize}[leftmargin = 0.5cm]
    \item Defining a quantitative hardness measurement is non-trivial since obtaining the optimal ground-truth solution of a TSP instance is NP-hard. Besides, as the TSP solver learns continuously during training, the hardness measurement must be updated adaptively. 
    \item Even equipped with a hardness measurement, we need a generative model to generate TSP instances with different hardness levels. In particular, generating sufficiently difficult samples is challenging.
    \item After obtaining instances with desired hardness, we need to fully utilize these instances to train more powerful TSP solvers, especially for the ability of generalizing to different hardness. 
\end{itemize}

To tackle these challenges, we first propose a principled hardness measurement to quantify the hardness of TSP instances. Specifically, we calculate the hardness as greedy self-improving potentials by comparing the current solver cost with a surrogate model. In this way, we avoid using the ground-truth optimal solution of TSP instances, calculating which is NP-hard and impractical. Besides, the hardness measurement is adaptive as the model learns continuously. Using the hardness measurement, we propose a hardness-adaptive generator, which can generate TSP instances with different hardness levels. Finally, we propose a curriculum learner to fully utilize the hardness-adaptive TSP instances generated by the generator. By learning instance weights, our method can train the TSP solvers more efficiently through curriculum learning. 

We conduct extensive experiments to verify the designs of our proposed method. Experimental results show that our hardness-adaptive generator can produce instances 10x harder than the existing methods, i.e., fixed uniform samples. Besides, our proposed curriculum learner, together with the hardness-adaptive generator, can achieve significant improvement over state-of-the-art models in terms of the optimality gap when training and test data have different distributions. The codes\footnote{https://github.com/wondergo2017/TSP-HAC. } are publicly available. 

In summary, we make the following contributions: 
\begin{itemize}[leftmargin = 0.4cm]
  \item We propose a principled hardness measurement using greedy self-improving potentials and surrogate models, avoiding the unbearable computational costs of calculating ground-truth optimal solution for TSP. 
  \item We design a hardness-adaptive generator to efficiently and continuously generate instances with different levels of hardness tailored for model training. 
  \item We propose a curriculum learner to fully utilize hardness-adaptive instances for more efficient model training with better performance.
\end{itemize}

\section{Related Work}
\subsection{Travelling Salesman Problem}
Recently, some works propose to use neural network models to facilitate solving TSP, including Pointer Networks~\cite{Ptr-RL-TSP17,Ptr-RL-VRP18,Ptr-TSP15}, Network Embedding~\cite{NE-RL-TSP17}, and Graph Neural Networks~\cite{TSP-paradigm19,GAT-RL-TSP19,GCN-SL-TSP18,GNN-Ptr-RL-TSP19,generalize-small-TSP20,generalization-TSP20,TSP-transformer21}. There are two paradigms for training TSP models: supervised learning and reinforcement learning(RL)~\cite{TSP-paradigm19}. Supervised methods adopt optimal routes as training labels, and the goal is to predict whether an edge exists in the optimal solution. Since ground-truth optimal solutions for TSP have to be used for training, this paradigm is not scalable for large-scale TSP instances. On the other hand, RL-based methods do not need optimal solutions of TSP. These methods commonly use a neural network encoder to obtain embeddings for each node, then use an RL controller to decide whether an edge is in the optimal solution step by step according to the state in RL. Once the output is obtained, the tour length acts as a negative reward for the RL controller to update the policy.
These models obtain impressive performance when training and testing data have the same distribution, e.g., Kool et al. \cite{GAT-RL-TSP19} achieves near-perfect results for TSP instances with 50 nodes on uniformly sampled instances.

\subsection{TSP Hardness and Size Generalization}
Previous learning-based TSP methods generate training and testing data from the same distribution, e.g., uniform samples from a unit square. In such a setting, it is unclear whether the proposed method actually learns to solve TSP instances or simply memorizes the training data.

Recent works~\cite{generalization-TSP20,generalize-small-TSP20} post similar questions by showing that the learned TSP solvers fail to generalize to different TSP sizes (i.e., the number of nodes in TSP) in training and testing. Our work takes a step forward by showing that even for the same size, different distributions can lead to diverse difficulties and result in poor generalization.

There are few heuristic hardness measurements for TSP instances. For example, Smith et al.~\cite{understand-hard-TSP10} define hardness-related instance features, which is closely related to the searching cost of heuristic solvers such as Lin–Kernighan (LK). Hemert~\cite{property-TSP05} defines the number of switching edges in LK as hardness. However, these measurements are based on non-learning-based TSP heuristics and are not suitable for learning-based solvers. In this paper, we define a principled hardness measurement for learning-based TSP solvers, which is also adaptive as the solver continuously learns.

\subsection{Curriculum Learning}
Curriculum learning~\cite{curriculum-survey-RL20,curriculum-survey-SL20} studies how to improve training strategies by manipulating the data distribution according to the model training stage, aiming to help the model train efficiently and obtain better performance~\cite{bengio2009curriculum,hacohen2019power}.

A family of classical methods trains the target model with samples from easy to challenging~\cite{bengio2009curriculum,kumar2010self,platanios2019competence,guo2018curriculumnet,florensa2017reverse}, mimicking how humans learn.
To perform curriculum learning, one has to first measure the hardness of samples and then use a curriculum scheduler to determine when and how to feed data into the model. 
In many fields, hardness measurement is based on domain knowledge (e.g., sentence length in machine translation~\cite{platanios2019competence}, signal to noise ratio in speech recognition~\cite{ranjan2017curriculum}, and others~\cite{tudor2016hard,soviany2020image,wei2016stc}).
For learning-based TSP, Lisicki et al.~\cite{curriculum-TSP20} takes the size of TSP as an indicator of hardness and train the model by gradually increasing the TSP size. In this paper, we focus on TSP instances of the same size but come from different distributions.

\section{Problem Formulation and Preliminary Study}
\subsection{Notations and Problem Formulation}
Following previous works, we focus on 2D-Eucliean TSP. Given the coordinate of $n$ nodes, $\mathbf{X}=\{\mathbf{x}_i\}^n_{i=1}$, 
where $\mathbf{x}_i \in [0,1]^2$ is the 2-D coordinate of node $i$ and the distance between node pairs $\mathcal{D}(\mathbf{x}_i,\mathbf{x}_j) = \left\|\mathbf{x}_i-\mathbf{x}_j\right\|_2$ is Euclidean, the objective is to find the shortest possible route that visits each node exactly once. The solution output by a TSP solver is a permutation of $n$ nodes $\bm{\pi} = \left[ \pi_1,\cdots,\pi_n \right]$  that minimizes the total tour length. Formally, the cost of solution $\bm{\pi}$ is defined as 
\begin{equation}
\mathcal{C} \left(\bm{\pi}, \mathbf{X} \right)=\left\| \mathbf{x}_{\pi_{n}}-\mathbf{x}_{\pi_{1}}\right\|_{2}+\sum \nolimits_{i=1}^{n-1}\left\| \mathbf{x}_{\pi_{i}}- \mathbf{x}_{\pi_{i+1}}\right\|_{2}.
\end{equation}
Each instance $\mathbf{X}$ has a minimal (i.e., optimal) solution cost $\mathcal{C}^*\left(\mathbf{X}\right) = \min_{\bm{\pi}} \mathcal{C} \left(\bm{\pi}, \mathbf{X} \right)$. Denote a TSP solver as $M$ and the cost of its solution as $\mathcal{C}_M\left( \mathbf{X}\right)$. 
Then $\mathcal{C}^*\left(\mathbf{X}\right) \leq \mathcal{C}_M\left( \mathbf{X}\right), \forall M$. 
Besides, no polynomial solver for TSP is known to date, i.e., an exact solver guaranteed to obtain the optimal solution for any TSP instance will evitably have an exponential time complexity, which is unbearable in practice. Learning-based TSP solvers aim to give near-optimal solutions while ensuring acceptable computational cost. We adopt the optimality gap as an optimization metric defined as:  
\begin{equation}
\mathcal{G}_M \left(\mathbf{X} \right)= \frac{\mathcal{C}_M\left( \mathbf{X}\right)-\mathcal{C}^*\left(\mathbf{X}\right)}{\mathcal{C}^*\left(\mathbf{X}\right)}.
\end{equation}
Note that we do not directly adopt $\mathcal{C}_M\left(\mathbf{X} \right)$ or $\mathcal{C}_M\left(\mathbf{X} \right)-\mathcal{C}^*\left(\mathbf{X}\right)$ because the optimal solution $\bm{\pi}^*$ remains the same when the distance of the instance is scaled, e.g., for $\mathbf{X}^\prime = a \mathbf{X}$, where $a > 0$ is a constant. But in those cases, $\mathcal{C}_M\left(\mathbf{X}^\prime \right) = a\mathcal{C}_M\left(\mathbf{X} \right) $ and thus is not comparable among instances with different scales. On the other hand, $\mathcal{G}_M (\mathbf{X}^\prime) = \mathcal{G}_M (\mathbf{X})$ and does not suffer from this problem.

For a dataset $\{\mathbf{X}_i\}^K_{i=1}$, where $K$ is the number of instances, the optimality gap is averaged for all the instances, i.e., $\mathcal{G}_M\left(\{\mathbf{X}_i\}^K_{i=1}\right)=\frac{1}{K}\sum_{i=1}^K \mathcal{G}_M\left(\mathbf{X}_i\right)$, unless stated otherwise.

\subsection{Preliminary Study}\label{sec:prelim}
To investigate whether existing current learning-based TSP solvers can generalize to different distributions, we first report preliminary study results. 

\subsubsection{Gaussian Mixture Generator}\label{sec:gm}
Following previous studies of TSP~\cite{understand-hard-TSP10}, Gaussian mixture generators can generate TSP instances with different hardness levels than the simple uniform sampling. Next, we introduce the Gaussian mixture generator in detail.

First, we sample the number of clusters $n_c$ from a discrete uniform distribution
\begin{equation}
    n_c \sim \mathcal{U} \{c_{\text{min}},\cdots,c_{\text{max}}\},
\end{equation}
where $c_{\text{min}}$ and $c_{\text{max}}$ are hyper-parameters. In our experiments, we set $c_{\text{min}} = 3$ and $c_{\text{max}} = 7$. Denote the cluster that node $i$ belongs to as $c_i$. Nodes have an equal probability of joining each cluster, i.e.,
\begin{equation}
    c_i \sim \mathcal{U} \{1,\cdots,n_c \}.
\end{equation}

Each cluster $i,1\leq i \leq n_c$, has a center vector $\bm{\mu}_i =\left(\mu_{i1},\mu_{i2} \right)$. The center vector is uniformly sampled in a square with length $c_{\text{DIST}}$, i.e.,
\begin{equation}
    \bm{\mu}_i \sim \mathcal{U} \left[0,c_{\text{DIST}}\right]^2.
\end{equation}
 The coordinate of node $i$, $\mathbf{x}_i$, is sampled from a Gaussian distribution $\mathbf{x}_i \sim \mathcal{N}(\bm{\mu}_{c_i},\mathbf{I})$. In this way, nodes belonging to the same cluster are close to each other.
To normalize the ranges of TSP instances, we follow previous works and scale the node coordinates $\mathbf{X} = \{\mathbf{x}_i\}_{i=1}^K$ into a unit square by using a min-max projection function $\phi\left(\mathbf{x}_i; \mathbf{X}\right)$ where $\phi(\cdot)$ is defined as: 
\begin{equation}\label{eq:project}
    \phi \left(\mathbf{x}_i;\mathbf{X}\right)= \frac{\mathbf{x}_i- \min(\mathbf{X})}{\max(\mathbf{X})-\min(\mathbf{X})},
\end{equation}
where $\min$ and $\max$ are calculated dimension-wise.

\subsection{Empirical Results}
Next, we present our experimental results. Specifically, following previous works, the TSP solver is trained using uniform sampling (more details of the solver can be found in Section~\ref{sec:expsetup}). Then, we test the solver using the Gaussian mixture generator.   

The results when fixing the adjustable parameter $c_{\text{DIST}}$ as 100 are shown in Figure~\ref{fig:weak measure}. Though the existing solvers show reasonably good results when the testing distribution is identical to training (i.e., uniform sampling), the performance drops significantly when using Gaussian mixture as the testing distribution. In fact, the maximum optimality gap even scales by more than 80 times. These results clearly demonstrate the shortcoming of the existing models when training and testing data have different distributions. 

Next, we show the results of varying $c_{\text{DIST}}$ in Table~\ref{tab:GMM hardness controller}. Empirically, we find that for the current learning-based TSP solvers, the optimality gap increases with $c_{\text{DIST}}$, i.e., instances with larger $c_{\text{DIST}}$ are more difficult. Since uniform sampling does not divide nodes into clusters, it is equivalent to $c_{\text{DIST}} = 0$ before projection function.
Therefore, as $c_{\text{DIST}}$ grows larger, the training and testing distribution become more diverse, and the weakness of the existing methods becomes more apparent.

\begin{table}
\centering

\small

\begin{tabular}{c|ccc}
\toprule
$c_{\text{DIST}}$    & 10 & 20 & 30  \\ 
Optimality Gap(\%) & $3.5\pm 0.2 $&$7.2\pm 0.4 $&$10.1\pm 0.6 $   \\ \midrule
$c_{\text{DIST}}$ & 50 & 70 & 100 \\ 
Optimality Gap(\%) &  $14.0\pm 1.0 $&$17.1\pm 1.5 $&$20.0\pm 1.6 $ \\
\bottomrule
\end{tabular}
\caption{The experimental results when varying $c_{\text{DIST}}$ for the Gaussian mixture generator.}
\label{tab:GMM hardness controller}
\end{table}

\section{Method}
\begin{figure*}%
    \centering
    \subfloat[\centering Generators ]{{\includegraphics[width=.5\columnwidth]{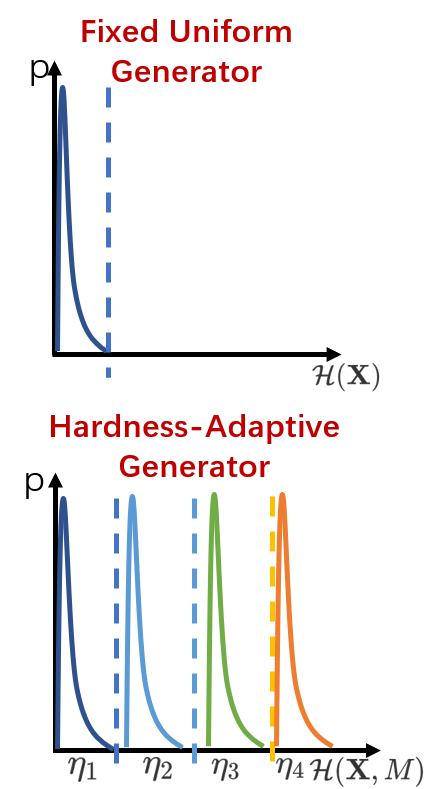} }}%
    \qquad
    \subfloat[\centering Our proposed hardness-adaptive curriculum learner]{{\includegraphics[width=1.2\columnwidth]{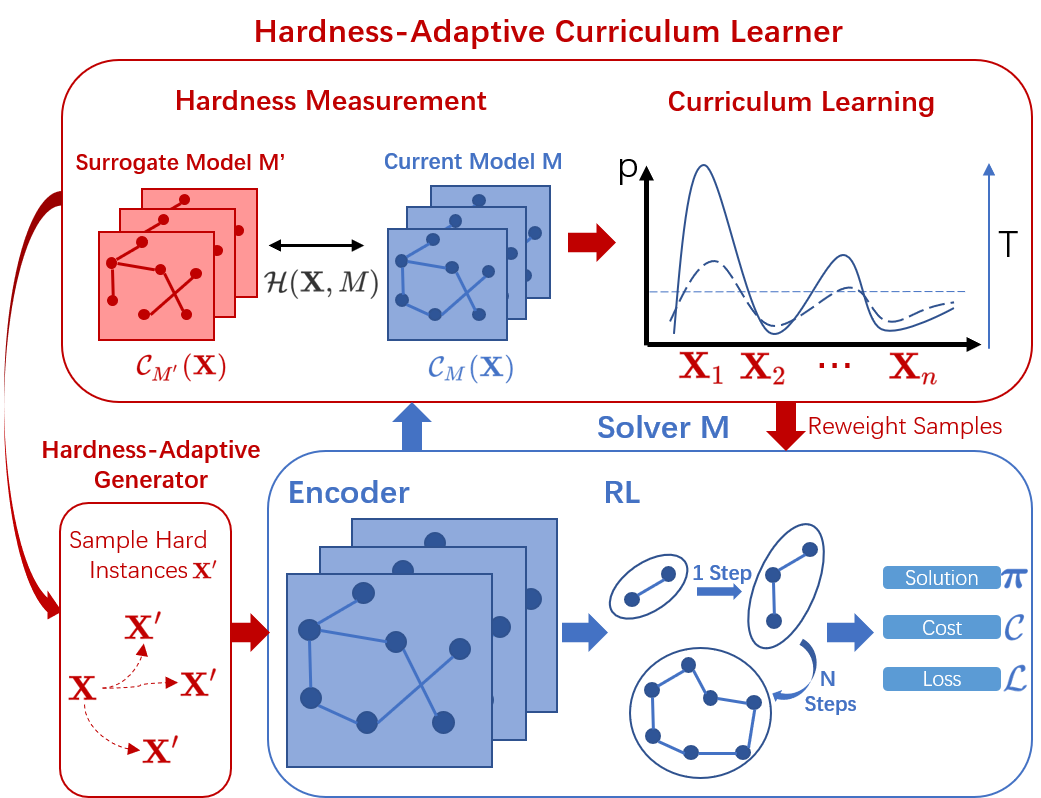} }}%
    \caption{(a) A comparison between the fixed uniform generator and our proposed hardness-adaptive generator. (b) The curriculum learner framework: Solver $M$ takes the generated samples and adopts an encoder and RL to calculate the cost. Information from the generator, samples, and the model is fed into the curriculum learner to exploit the current training stage, reweight samples, and update model parameters. The generator continues to generate hardness-adaptive samples for the current model to keep improving model performance.(Best viewed in color)}
    
    \label{fig:framework}
\end{figure*}

In this section, we introduce our proposed method. First, we introduce how to measure the hardness of TSP instances. Then, we propose a hardness-adaptive generator to sample hardness-diverse TSP instances. Lastly, we introduce a curriculum learning model to fully utilize these samples and better train the solver. The overall framework of our method is shown in Figure~\ref{fig:framework}.

\subsection{Hardness Measurement}
To enable TSP solvers to handle various distributions, we need instances with different hardness levels. Therefore, we need a quantitative hardness measurement. Denote the hardness of instance $\mathbf{X}$ to solver $M$ as $\mathcal{H}(\mathbf{X},M)$. Notice that the hardness is solver-specific since the hardness can change as the solver is trained.

 Some previous work~\cite{curriculum-TSP20} takes the TSP size $n$ as an indicator of hardness since large-size instances are usually more complex and difficult to solve. In this paper, we take an orthogonal direction by considering the hardness of instances with the same size but having different distributions. 
 
 A straightforward idea is to use the optimality gap $\mathcal{G}_M(\mathbf{X})$ as the hardness since it naturally measures the quality of the solver compared to the optimal solution. However, it is impractical to obtain the optimal solution cost $\mathcal{C}^*\left(\mathbf{X}\right)$ since itself is another NP-hard problem, i.e., the TSP decision problem. Therefore, we need to directly calculate $\mathcal{H}(\mathbf{X},M)$ without calculating $\mathcal{C}^*\left(\mathbf{X}\right)$. 

Another challenge is that the solver $M$ can vary greatly during training. For example, as the solver learns, instances considered hard in previous stages may not be hard anymore in the current stage, and it is crucial for the hardness measurement to capture such a drift. 
To solve this problem, we propose a hardness measurement using self-improving potentials
\begin{equation}\label{eq:hardness}
\mathcal{H}(\mathbf{X},M)=  \frac{\mathcal{C}_M(\mathbf{X})-\mathcal{C}_{M^\prime}(\mathbf{X})}{\mathcal{C}_{M^\prime}(\mathbf{X})},
\end{equation}
where $M^\prime$ is a greedily updated surrogate model, e.g., when $M$ is differentiable, we can conduct several steps of gradient descends to obtain $M^\prime$. Intuitively, instead of comparing with the optimal solution, we compare $M$ with a potential solution as the surrogate, i.e., if we can find $M^\prime$ with a much smaller cost than $M$, it means $\mathbf{X}$ is still ``hard'' for $M$. From another perspective, the hardness measurement is also a lower bound of the ground-truth optimality gap

\begin{equation}
\begin{aligned}
\mathcal{H}(\mathbf{X},M)&=  \frac{\mathcal{C}_M(\mathbf{X})-\mathcal{C}_{M^\prime}(\mathbf{X})}{\mathcal{C}_{M^\prime}(\mathbf{X})} 
\\ &\leq  \frac{\mathcal{C}_M\left( \mathbf{X}\right)-\mathcal{C}^*\left(\mathbf{X}\right)}{\mathcal{C}^*\left(\mathbf{X}\right)} = \mathcal{G}_M \left( \mathbf{X}\right).
\end{aligned}
\end{equation}
The equality holds when our surrogate model $M^\prime$ produces the optimal solution for $\mathbf{X}$.

Notice that if solver $M$ outputs routes with probabilities rather than being deterministic, we define cost function as an expectation, i.e.,$\mathcal{C}_M(\mathbf{X}) = \mathbb{E}_{ p_M(\bm{\pi} | \mathbf{X})} \left[\mathcal{C}(\bm{\pi},\mathbf{X}) \right] $,
where $p_M(\bm{\pi} | \mathbf{X})$ is the probability of solver $M$ outputing route $\bm{\pi}$ for instance $\mathbf{X}$. We do not adopt minimum, i.e.,  $\mathcal{C}_M(\mathbf{X}) = \min_{\bm{\pi}} \left[ \mathcal{C}(\bm{\pi},\mathbf{X}), \bm{\pi} \in M \right]$ because such measurement is difficult to be optimized due to differentiability problems. Therefore, the hardness measurement is also defined as an expectation. As enumerating all possible routes is intractable, we sample routes to obtain an unbiased estimator.

\subsection{Hardness-Adaptive Generator}
As shown in Section~\ref{sec:prelim}, naive i.i.d. sampling can not produce sufficiently difficult TSP instances. Next, we introduce a hardness-adaptive generator to sample hardness-diverse TSP instances using the hardness measurement $\mathcal{H}(\mathbf{X},M)$.

The main difficulty of designing a hardness-adaptive generator is to generate sufficiently difficult samples. To solve that challenge, 
inspired by energy-based models~\cite{EBM21}, we adopt an energy function as the generative model. Specifically, we define the energy function using hardness as $E(\mathbf{X}|M)=-\mathcal{H}\left(\mathbf{X},M \right)$. Then, the probability distribution is defined as
\begin{equation}
\begin{aligned}
    p\left(\mathbf{X} | M \right)&= \frac{\exp \left( -E(\mathbf{X}) | M\right)}{\int_{\mathbf{X}^\prime} \exp \left(-E(\mathbf{X}^\prime)|M \right)\mathrm{d} \mathbf{X}^\prime}  \\ &= \frac{\exp(\mathcal{H}\left( \mathbf{X},M \right))}{\int_{\mathbf{X}^\prime} \exp \left( \mathcal{H}\left( \mathbf{X}^\prime,M \right) \right)\mathrm{d} \mathbf{X}^\prime}.
\end{aligned}
\end{equation}
In other words, instances with larger $\mathcal{H}\left( \mathbf{X},M \right)$ are more likely to be sampled.
When generating the samples, we first randomly generate an instance $\mathbf{X}^{(0)}$, e.g., using uniform sampling. Then, inspired by Langevin Dynamics~\cite{langevin19}, we further optimize the sample to increase the hardness by doing gradient ascend

\begin{equation}\label{eq:dynamic}
\begin{aligned}
\mathbf{X}^{(t)'}&=\mathbf{X}^{(t)}+\eta \nabla_{\mathbf{X}^{(t)}} \log p(\mathbf{X}^{(t)}) \\ 
&=\mathbf{X}^{(t)}+\eta \nabla_{\mathbf{X}^{(t)}} \mathcal{H}(\mathbf{X}^{(t)},M), \\ 
\mathbf{X}^{(t+1)}&=\phi(\mathbf{X}^{(t)'}),
\end{aligned}
\end{equation}

where $t$ denotes the number of optimization steps, $\phi(\cdot)$ is the projection function in Eq.~\eqref{eq:project} to ensure the validity of the generated samples, and $\eta$ is the gradient step size. 
In the generator, we can adjust $t$ and $\eta$ to generate instances with different hardness.

For methods outputting solutions with probabilities, directly calculating the gradient in Eq.~\eqref{eq:dynamic} also needs to enumerate all possible solutions $\boldsymbol{\pi}$, which is intractable. Using policy gradient, we have

\begin{small}
\begin{equation}\label{eq:hardnessgradient}
\begin{aligned}
 &\nabla_{\mathbf{X}} \mathcal{H}(\mathbf{X},M) \\ &=\mathbb{E}_{p_{M(\boldsymbol{\pi}\mid \mathbf{X})}}
 \left[ \frac{\mathcal{C}_M(\mathbf{X})}{\mathcal{C}_{M^\prime}(\mathbf{X})} \nabla_\mathbf{X} \log p_M(\boldsymbol{\pi}\mid \mathbf{X})  + \frac{\nabla_\mathbf{X} \mathcal{C}_M(\mathbf{X}) }{\mathcal{C}_{M^\prime}(\mathbf{X})}\right].
\end{aligned}
\end{equation}
\end{small}

In deriving the above equation, we assume $\mathcal{C}_{M^\prime}(\mathbf{X})$ to be a constant.
Then, we can sample routes from $p_M\left(\bm{\pi} | \mathbf{X} \right)$ to obtain unbiased estimators of Eq.~\eqref{eq:hardnessgradient}. 

\begin{algorithm}[tb]
\caption{Hardness-adaptive Curriculum Learner}         
\label{algo: framework}
\begin{algorithmic}[1]
\REQUIRE 
Hardness-adaptive and uniform generator, TSP solver $M$, batch size $B$, training epochs $L$

\STATE Initialize and warm up $M$ using uniform sampling 
\FOR{$l = 1, \dots, L$}

    \STATE Randomly generate dataset $D$ by uniform samples and hard samples using Eq.~\eqref{eq:dynamic}
    \FOR{$b = 1, \dots, |D|/B$}
        \STATE Get batch data $\{\mathbf{X}\}_{i=1}^B$ from $D$  \\
        \STATE Calculate hardness $\mathcal{H}(\mathbf{X},M)$ for $\{\mathbf{X}\}_{i=1}^B$ using Eq.~\eqref{eq:hardness} 
        \STATE Calculate sample weights $w_i$ using Eq.~\eqref{eq:sampleweights} 
        \STATE Calculate gradients for each instance using Eq.~\eqref{eq:solvergradient}
  
        \STATE Update model parameters using weighted gradients $\nabla \mathcal{L}_{\bm{\theta}}\left(\{\mathbf{X}_i\}_{i=1}^B \right)=\sum \nolimits_i w_i \nabla \mathcal{L}_{\bm{\theta}}(\mathbf{X}_{i})$ %
    \ENDFOR
    \STATE Increase the curriculum learner temperature $T$
\ENDFOR
\end{algorithmic}

\end{algorithm}

\begin{table*}[h]
\centering

\begin{tabular}{l|lll}
\toprule
Testing Distribution 
 & Model Cost & Optimal Cost  & Optimality Gap (\%) \\ \midrule
Uniform Generator & $5.80\pm 0.01 $&$ 5.70\pm 0.01$&$ 1.70\pm 0.05$    \\ 
Hardness-adaptive Generator($\eta$=5) & $4.86\pm 0.05 $&$ 4.35\pm 0.04$&$ 12.36\pm 1.21$     \\ \bottomrule
\end{tabular}
\caption{
The results of different generators using a TSP solver pre-trained on uniform sampling
}
\label{tab:hard sample generation}
\end{table*}

\begin{figure*}[h]%
    \centering
    \subfloat[Samples generated by uniform generator ]{{\includegraphics[width=6cm]{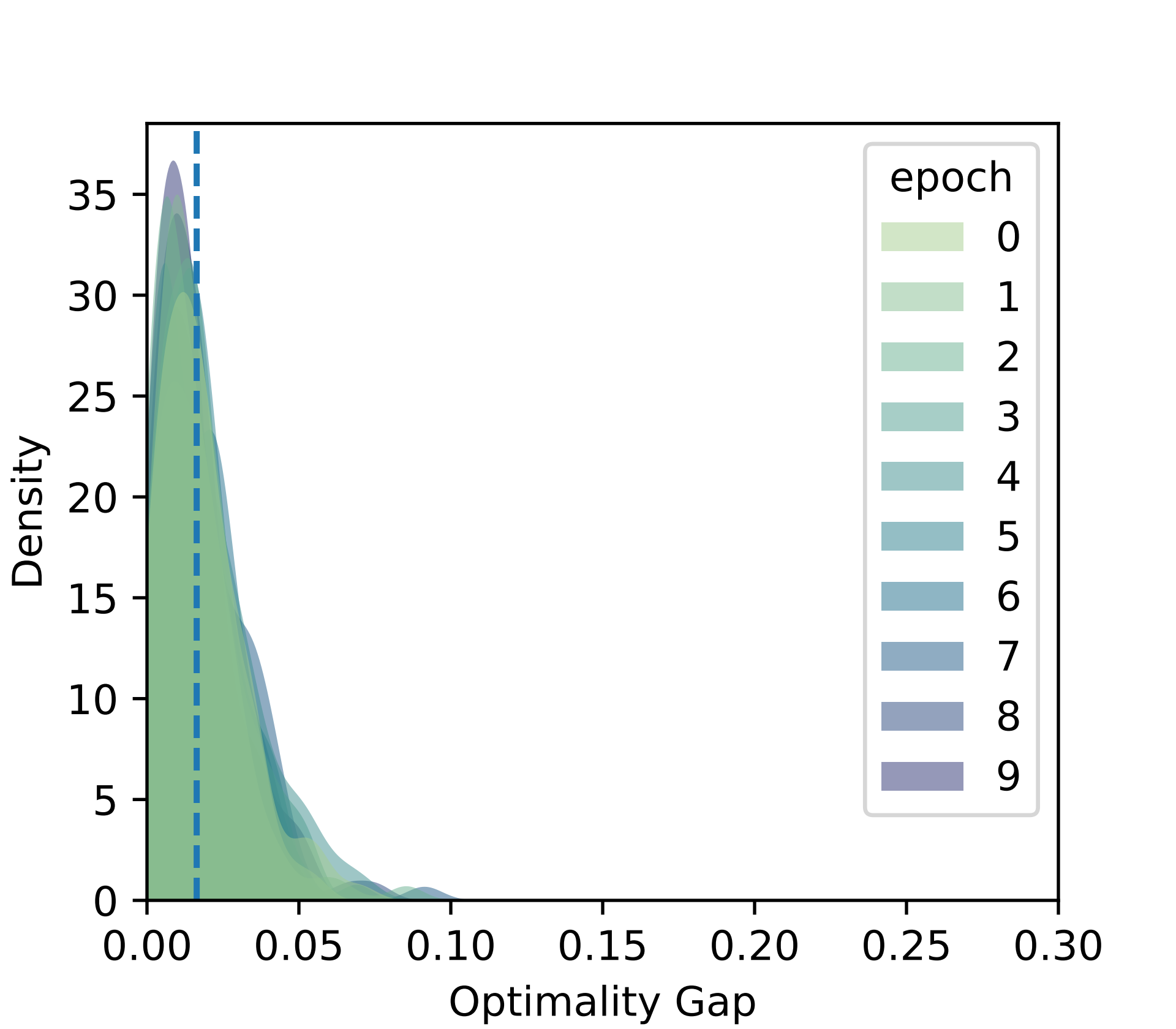} }}%
    \qquad
    \subfloat[Samples generated by our hardness-adaptive generator ($\eta=5$). ]{{\includegraphics[width=6cm]{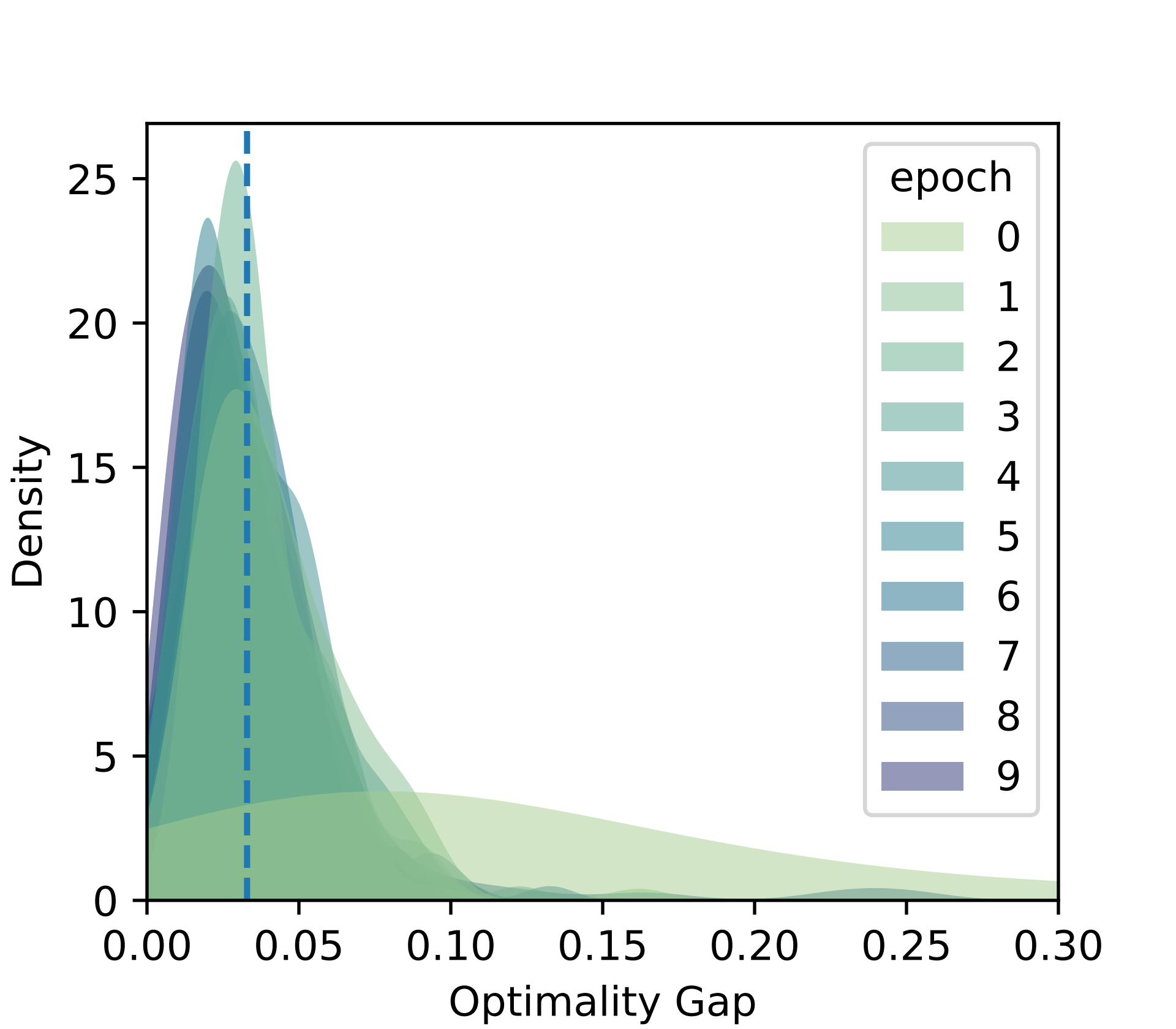} }}
    \caption{The results of comparing the optimality gap of TSP instances generated by a uniform generator and our proposed hardness-adaptive generator. Our proposed generator can continuously generate instances with more diverse hardness levels. (Best viewed in color)} 
    \label{fig:gen-train}%
\end{figure*}

\subsection{Hardness-Adaptive Curriculum Learner}\label{sec:cl}
To learn a more powerful TSP solver, we explore hardness-adaptive samples through curriculum learning. In this paper, we adopt a simple reweighting-based curriculum strategy, but our method can be straightforwardly generalized to more advanced curriculum learning methods. 

Consider a mixed dataset $\{\mathbf{X}_i \}_{i=1}^K$ with both hard and easy samples, e.g., generated by our hardness-adaptive generator and the fixed uniform generator. Intuitively, we should focus more on hard samples during training since they reflect the shortcomings of the current solver. Therefore, we propose to reweight TSP instance $\mathbf{X}_i$ using the hardness as follows,
\begin{equation}\label{eq:sampleweights}
\small
    w_i=\frac{\exp(\mathcal{F}(\mathcal{H}(\mathbf{X}_i,M))/T)}{\sum_j \exp(\mathcal{F}(\mathcal{H}(\mathbf{X}_j,M))/T)},
\end{equation}
where $\mathcal{F}(\cdot)$ is a transformation function and $T$ is a temperature to control the stage of the curriculum. When the temperature is high, all the samples are treated roughly equally. As the temperature $T$ goes down, the weight distribution shifts and  harder samples are assigned larger weights than easy samples. For a data batch $\{\mathbf{X}_i\}_{i=1}^B$, the reweighted loss is calculated as follows
\begin{equation}\label{eq:weightedloss}
\small
    \mathcal{L}_{\bm{\theta}}\left(\{\mathbf{X}_i\}_{i=1}^B \right)=\sum \nolimits_i w_i\mathcal{L}_{\bm{\theta}}(\mathbf{X}_{i}),
\end{equation}
where $\bm{\theta}$ denotes the learnable parameters of the TSP solver. 
Following previous works~\cite{GAT-RL-TSP19}, we use reinforcement learning to optimize the TSP solver.
The loss function is defined as 
\begin{equation}
\mathcal{L}_{\bm{\theta}}(\mathbf{X})=\mathbb{E}_{p_{M_{\bm{\theta}}}(\boldsymbol{\pi} \mid \mathbf{X})} \left[\mathcal{C}\left(\boldsymbol{\pi} | \mathbf{X} \right) \right].
\end{equation}

Then we can use policy gradient and REINFORCE~\cite{REINFORCE92} algorithm to minimize the loss 
\begin{small}
\begin{equation}\label{eq:solvergradient}
    \nabla \mathcal{L}_{\bm{\theta}} ( \mathbf{X})=\mathbb{E}_{p_{M_{\boldsymbol{\theta}}}(\boldsymbol{\pi} \mid \mathbf{X})}\left[ \left(\mathcal{C}(\boldsymbol{\pi}| \mathbf{X})-\mathcal{C}_b(\mathbf{X}) \right) \nabla \log p_{M_{\boldsymbol{\theta}}}(\boldsymbol{\pi} \mid \mathbf{X})\right],
\end{equation}
\end{small}
where $\mathcal{C}_b(\mathbf{X})$ is the cost of a baseline to reduce gradient variances. Following~\cite{GAT-RL-TSP19}, we set $\mathcal{C}_b(\mathbf{X})$ as a deterministic greedy rollout of the best model so far.
The pseudo code is shown in Algorithm~\ref{algo: framework}.

\section{Experiments}
In this section, we conduct experiments 
to answer the following questions:
\begin{itemize}[leftmargin = 0.5cm]
    \item \textbf{Q1}: Can our hardness-adaptive generator continuously  generate samples with diverse hardness levels than uniform sampling?
    \item \textbf{Q2}: Can our curriculum learning model train more powerful TSP solvers that are better generalizable to different distributions?
\end{itemize}

\begin{table*}[h]
\centering

\begin{tabular}{l|lllll}
\toprule
$\eta$                   & 0.1 & 0.5 & 1 & 5 & 10  \\ \midrule
Optimality Gap (\%) & $0.7\pm 0.1 $&$1.5\pm 0.2 $&$2.9\pm 0.4 $&$12.4\pm 1.2 $&$14.7\pm 1.4 $ \\ \bottomrule
\end{tabular}
\caption{Relationship between the optimality gap and $\eta$ for our hardness-adaptive generator.}
\label{tab:AGConroller}
\end{table*}

\subsection{Experimental Setup}\label{sec:expsetup}
In our experiments, we focus on TSP-50, i.e., instances with 50 nodes. 
Throughout our experiments, we adopt GAT~\cite{GAT-RL-TSP19} as the encoder in our TSP solver with all hyperparameters are kept the same as in the original paper. 
Besides, we adopt Gurobi~\footnote{https://www.gurobi.com}, a dedicated non-learning-based TSP solver, to obtain the optimal cost $\mathcal{C}^*$ as ground-truths.

\subsection{Experimental Details}
For our experimental settings and model training, we largely follow \cite{GAT-RL-TSP19} except that the TSP instances are generated using different methods. For experiments in Section 5.3, we sample 100,000 TSP instances for each training epoch from the uniform generator or our hardness-adaptive generator ($\eta=5$). Then, we sample 100 randomly selected instances as the testing set and calculate the average optimality gap and standard deviation.
For experiments in Section 5.4, we sample 10,000 instances for each training epoch. The testing dataset is composed of 10,000 instances generated by the Gaussian mixture generator. For the TSP solver, the GAT has three encoding layers with the embedding dimensionality 128 and the number of attention heads 8. Training methods such as batch normalization, tanh clipping of action logits and clipping of gradient norms are kept unchanged as in the original paper to stabilize the training process. We use Adam optimizer with the learning rate of 0.0001, the weight decay of 1.0, and the batch size of 512. The significance threshold used to update the RL baseline is $\alpha=0.05$.

\begin{figure}[h]
    \centering    \includegraphics[width=1\columnwidth]{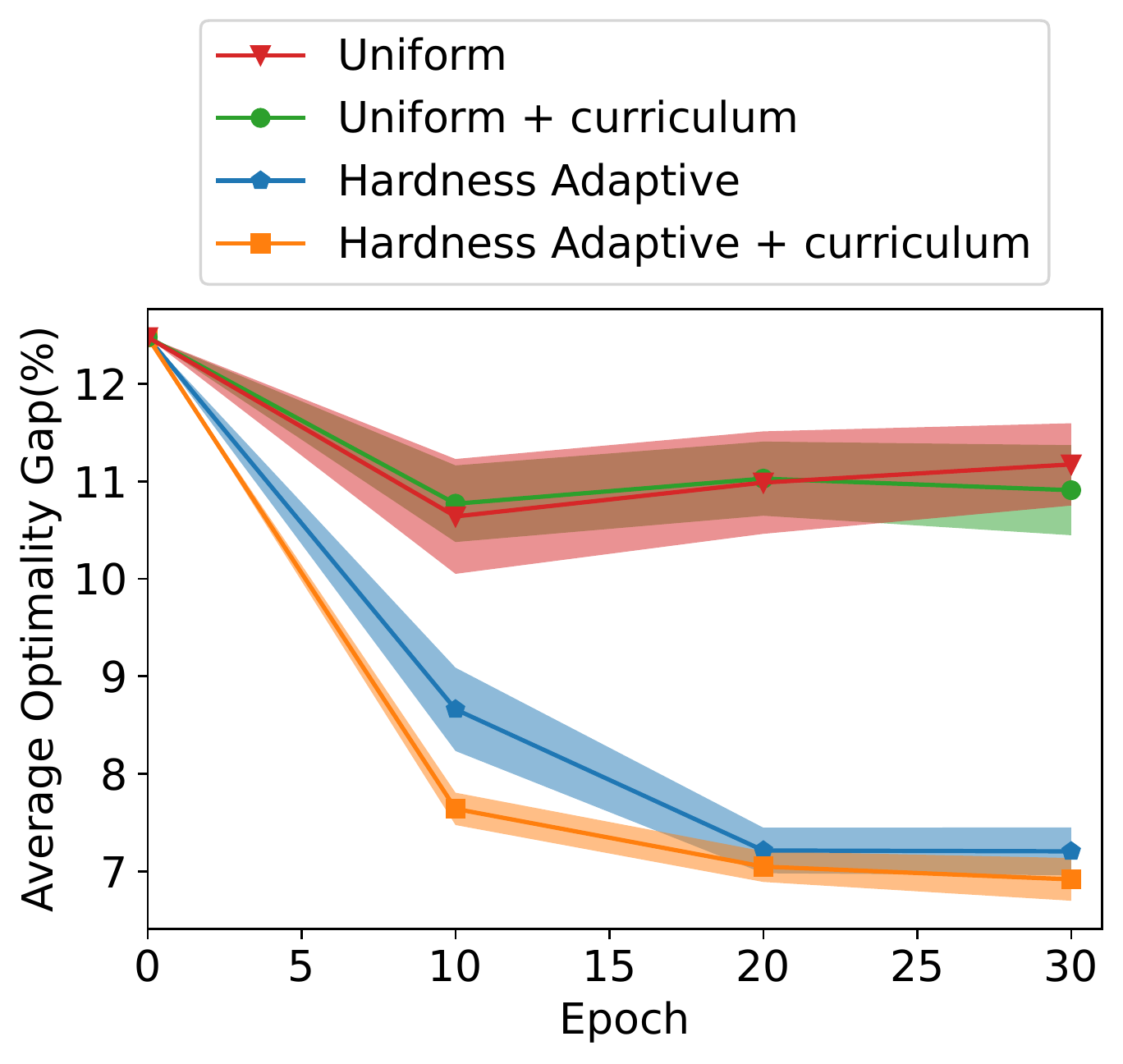}
    
    \caption{The experimental results when the training distribution is uniform and the testing distribution is Gaussian mixture (unknown to the model). Our proposed hardness-adaptive generator and curriculum learner significantly reduces the optimality gap and converges faster.}
    \label{fig:curriculum-perform}
\end{figure}

\subsection{Generating Hardness-Adaptive Instances}\label{sec:exp:generate}
To answer \textbf{Q1}, we compare the hardness distribution of instances sampled by different generators. The TSP solver is pre-trained on 
samples from the uniform distribution. 

First, we use a uniform generator and our proposed hardness-adaptive generator to generate instances and test the pre-trained solver. The results are shown in Table~\ref{tab:hard sample generation}. Our proposed hardness-adaptive generator can generate more difficult TSP instances, demonstrating its effectiveness. These instances can be utilized to further improve our TSP solver.  

Next, to test whether the generators can continuously generate hardness-adaptive samples as the solver learns, we fine-tune the solver using these generators, i.e., further optimize the solver using generators to generate training instances. Figure~\ref{fig:gen-train} shows the optimality gap (i.e., ground-truth hardness) distribution with different training epochs. As the epoch increases, samples generated by both generators tend to have a smaller optimality gap, but our proposed hardness-adaptive generator continuously generates instances with  more  diverse hardness levels.

Lastly, to verify the design of our hardness-adaptive generator that using a larger step size $\eta$ can generate more difficult samples, we report the results of varying $\eta$ in Table~\ref{tab:AGConroller}. As $\eta$ grows larger, the optimality gap increases, indicating that the generated instance is more difficult. The results are consistent with our design.

\subsection{Results when Training and Testing Distributions are Different}
To answer \textbf{Q2}, we conduct experiments when training and testing TSP instances are from different distributions. Specifically, we consider two data generators for the training data as Section~\ref{sec:exp:generate}, i.e., a uniform generator and our hardness-adaptive generator. For hardness-adaptive generator, we set $\eta=5$. Besides, we also compare two training paradigms: one using our proposed curriculum learner in Section~\ref{sec:cl} and another not using curriculum learning, i.e., all TSP instances have the same weights. For the testing data, we generate TSP instances using the Gaussian mixture generator introduced in Section~\ref{sec:gm}. Notice that in all the scenarios, the testing distribution is unknown to the model. We repeat the experiments 15 times and calculate the average optimality gap evaluated on the testing dataset as the final results.

As shown in Figure~\ref{fig:curriculum-perform}, models trained using our hardness-adaptive generator significantly reduce the optimality gap, which demonstrates that our generated hardness-adaptive samples can greatly improve the TSP solver when training and testing data come from different distributions. Besides, the figure also shows that our proposed curriculum learner helps reduce variance and makes the training process faster. 

\section{Conclusion}
In this paper, we explore whether the learning-based TSP solvers can generalize to different distributions. We propose a quantitative hardness measurement for TSP, a hardness-adaptive generator to generate instances with different hardness levels, and a curriculum learner to train the TSP solver. Experiments demonstrate the effectiveness of our proposed method in generating hardness-adaptive TSP instances and training more powerful TSP solvers when training and testing data have different distributions. For the future, it will deserve further investigations to extend our proposed hardness-adaptive curriculum learner to other combinatorial problems beyond the travelling salesman problem. We will also consider different hardness approximation methods, extention for non-differentiable solvers and the setting of continual learning in future works.

\section*{Acknowledgments}

This work was supported in part by the National Key Research and Development Program of China No. 2020AAA0106300 and National Natural Science Foundation of China No. 62102222 and Tsinghua GuoQiang Research Center Grant 2020GQG1014. All opinions, findings, conclusions and recommendations in this paper are those of the authors and do not necessarily reflect the views of the funding agencies.


\bibliography{aaai22.bib}
\end{document}